\documentclass[11pt,a4paper]{article}

\usepackage[utf8]{inputenc}
\usepackage[T2A,T1]{fontenc}
\usepackage[ukrainian,english]{babel}
\usepackage{amsmath,amssymb}
\usepackage{graphicx}
\usepackage{booktabs}
\usepackage{multirow}
\usepackage{hyperref}
\usepackage{xcolor}
\usepackage{natbib}
\usepackage{url}
\usepackage{microtype}
\usepackage{enumitem}
\usepackage{caption}
\usepackage{subcaption}
\usepackage[margin=1in]{geometry}
\usepackage{float}
\usepackage{array}
\usepackage{tabularx}
\usepackage{pgfplots}
\usepackage{pgfplotstable}
\pgfplotsset{compat=1.18}
\usepgfplotslibrary{colorbrewer}

\hypersetup{
    colorlinks=true,
    linkcolor=blue!70!black,
    citecolor=blue!70!black,
    urlcolor=blue!70!black,
}

\newcolumntype{R}[1]{>{\raggedleft\arraybackslash}p{#1}}

\title{The Tokenizer Tax Across 24 European Languages:\\
Domain Invariance, Cross-Lingual Few-Shot Effects,\\
and the Ukrainian Penalty}

\author{
  Volodymyr Ovcharov\thanks{Corresponding author: \texttt{volodymyr@legal.org.ua}} \\
  LEX AI Platform, legal.org.ua \\
  Kyiv, Ukraine
}

\date{May 2026}

\begin{document}

\maketitle

\begin{abstract}
Tokenizer fertility -- the number of tokens per word -- imposes a hidden cost on non-English NLP. We measure fertility for ten foundation models across 25~European languages on parallel text, producing the first controlled tokenizer tax map for the continent. The tax spans 2.5$\times$ from English (1.2 tokens/word) to Greek/Maltese (${\sim}$3.1), following a clear hierarchy: Romance (1.5--1.7), Germanic (1.7--1.9), Slavic (2.2--2.5), Uralic/Baltic (2.7--3.0). Ukrainian (2.7) pays 15--18\% more than cognate Slavic languages, reflecting underrepresentation in pre-training data. Fertility rankings are domain-invariant across three text registers ($\rho > 0.97$). A subword analysis reveals that high-fertility tokenizers fragment morphological boundaries rather than preserving them. Cross-lingual few-shot evaluation on four Slavic languages shows that few-shot effects are model-intrinsic, not language-dependent. We release all measurements as a public dataset.
\end{abstract}

\noindent\textbf{Keywords:} tokenizer fertility, language tax, cross-lingual evaluation, few-shot prompting, morphologically rich languages, Ukrainian NLP, Slavic languages

\section{Introduction}

Every token costs money. When a tokenizer fragments Ukrainian text into 2.7 tokens per word but handles English at 1.2, the same API call costs more than twice as much -- a hidden ``tokenizer tax'' that penalizes billions of speakers of morphologically rich languages. \citet{petrov2024language} and \citet{ahia2023all} documented this disparity, but existing measurements cover individual languages on heterogeneous text, making it impossible to disentangle language effects from content effects.

This paper provides four things the literature lacks. First, a \textbf{controlled cross-lingual fertility map}: we measure ten foundation models on parallel text in 25~European languages, isolating the tokenizer's contribution from content variation. Second, a \textbf{domain invariance test}: we compare fertility across three text registers -- legal, news, and encyclopedic -- to determine whether a single measurement suffices across applications. Third, a \textbf{subword analysis} that reveals how different tokenizers fragment morphologically complex Ukrainian words, explaining the fertility gap mechanistically. Fourth, a \textbf{downstream consequence test}: we evaluate few-shot classification across four Slavic languages to determine whether the tokenizer tax affects task performance.

\citet{ovcharov2026tokenizer} showed that tokenizer fertility varies by 1.6$\times$ across models on Ukrainian legal text and that few-shot prompting can degrade performance by up to 26~pp. We extend these findings along the cross-lingual axis and add three new model families (GPT-4o, Gemma~2, DeepSeek~V3):

\begin{enumerate}[leftmargin=*]
    \item \textbf{Domain invariance} (Experiment~1): Fertility rankings are stable across three text registers -- legal, news, and encyclopedic -- with Spearman $\rho > 0.97$ between domains. A single measurement predicts cost across all applications.

    \item \textbf{Cross-lingual fertility map} (Experiment~2): Fertility across 25~EU languages on parallel text, revealing a 2.5$\times$ spread from English (1.23) to Greek/Maltese (3.1). Ukrainian (2.66) pays 15--18\% more than cognate Slavic languages.

    \item \textbf{Subword analysis} (Experiment~2b): Morphological decomposition analysis showing that high-fertility tokenizers split at arbitrary byte boundaries rather than morpheme boundaries, explaining the performance gap.

    \item \textbf{Cross-lingual few-shot} (Experiment~3): Few-shot classification on SIB-200 across Ukrainian, Polish, Russian, and Czech. The effect is model-intrinsic: the same models that benefit in one language benefit in all four.

    \item \textbf{Linguistic competence} (Experiment~4): ULP benchmark (347 grammar questions) tests whether fertility predicts grammatical accuracy.
\end{enumerate}

\section{Related Work}

\subsection{Few-Shot Learning and Its Limits}

\citet{brown2020language} established few-shot in-context learning as a core capability of large language models. Subsequent work has shown that few-shot performance depends on example selection \citep{liu2022makes}, format consistency \citep{min2022rethinking}, and label distribution \citep{zhao2021calibrate}. \citet{min2022rethinking} demonstrated that even random labels can improve performance, suggesting that demonstrations primarily specify task format rather than input--output mappings.

However, these studies focus overwhelmingly on English. \citet{lai2023chatgpt} found significant performance variation across languages but did not systematically investigate few-shot effects. \citet{ovcharov2026tokenizer} documented systematic few-shot degradation on Ukrainian legal text, showing that the effect depends on model architecture rather than input language. Our work extends the cross-lingual dimension: testing whether the same patterns hold across four Slavic languages and 25~European languages.

\subsection{Tokenizer Fertility and Multilingual Fairness}

\citet{petrov2024language} formalized the ``language tax'' imposed by suboptimal tokenization, showing 2--15$\times$ token cost variation across languages. \citet{ahia2023all} demonstrated that API cost varies by an order of magnitude across languages for equivalent content. \citet{rust2021good} showed that monolingual performance correlates with pre-training data proportion, with fertility as a proxy.

These studies examine general-domain text or single languages. \citet{niklaus2024lextreme} introduced LEXTREME, a multi-lingual legal benchmark, but did not measure tokenizer fertility or few-shot effects. Alternative approaches bypass the tokenizer entirely: CANINE \citep{clark2022canine} operates on Unicode code points and ByT5 \citep{xue2022byt5} on raw bytes, eliminating the fertility disparity at the cost of longer sequences. \citet{zheng2021allocating} showed that vocabulary reallocation can reduce fertility by 15--30\% for underserved languages. We provide the first controlled cross-lingual comparison of six tokenizers across 25~languages on parallel text, isolating the tokenizer's contribution from content variation, and test domain invariance across three text registers.

\subsection{Morphologically Rich Languages in NLP}

Ukrainian, Polish, Russian, and Czech are Slavic languages with rich inflectional morphology: 7~cases, grammatical gender, and extensive verb conjugation. This morphological complexity interacts with subword tokenizers, potentially producing more fragmented representations that interfere with in-context pattern matching. \citet{conneau2020unsupervised} showed that multilingual model performance correlates with pre-training data volume, with low-resource languages suffering disproportionately. \citet{chaplynskyi2023ukrbruk} showed consistent underperformance of multilingual models on Ukrainian compared to English, a finding our cross-lingual experiments extend to the few-shot setting.

\section{Methodology}

\subsection{Models}

We extend the seven API models from \citet{ovcharov2026tokenizer} with three additional models whose tokenizers are publicly available, bringing the total to ten (Table~\ref{tab:models}). The original seven are accessed via the AWS Bedrock API; the three additions are measured using their HuggingFace tokenizers and OpenAI's \texttt{tiktoken} library. We validate the local tokenizer approach by comparing it against API measurements on Ukrainian news text for the three models available in both settings (Table~\ref{tab:validation}). Local and API fertility values agree within 1.8\% (mean absolute difference 0.030 tokens/word), confirming that local tokenizer measurement is a reliable and cost-free substitute.

\begin{table}[t]
\centering
\caption{Validation: API-reported vs.\ local tokenizer fertility on Ukrainian SIB-200 news text. $\Delta$ = absolute difference.}
\label{tab:validation}
\begin{tabular}{lR{1.2cm}R{1.2cm}R{1.0cm}R{1.0cm}}
\toprule
\textbf{Model} & \textbf{API} & \textbf{Local} & \textbf{$\Delta$} & \textbf{\%} \\
\midrule
Llama 4 Maverick & 2.197 & 2.158 & 0.039 & 1.8 \\
Llama 3.3 70B    & 2.333 & 2.317 & 0.016 & 0.7 \\
Qwen3 32B        & 3.583 & 3.617 & 0.034 & 0.9 \\
\midrule
\textit{Mean}    &       &       & \textit{0.030} & \textit{1.1} \\
\bottomrule
\end{tabular}
\end{table}

\begin{table}[t]
\centering
\caption{Models evaluated. Original seven via AWS Bedrock (April--May 2026); three additions (\dag) via local tokenizer. MoE = mixture of experts; active = parameters per forward pass.}
\label{tab:models}
\begin{tabular}{lllR{1.0cm}}
\toprule
\textbf{Model} & \textbf{Provider} & \textbf{Tokenizer} & \textbf{Vocab} \\
\midrule
Llama 4 Maverick   & Meta      & SentencePiece   & 200K \\
Llama 3.3 70B      & Meta      & SentencePiece   & 128K \\
Mistral Large 3    & Mistral   & SentencePiece   & 32K \\
Nemotron Super 3   & NVIDIA    & SentencePiece   & 128K \\
Nova Pro           & Amazon    & Proprietary     & --- \\
Qwen3 32B          & Qwen      & tiktoken-derived & 151K \\
GPT-4o\dag         & OpenAI    & tiktoken (o200k) & 200K \\
Gemma 2 27B\dag    & Google    & SentencePiece   & 256K \\
DeepSeek V3\dag    & DeepSeek  & Custom BPE      & 128K \\
\bottomrule
\end{tabular}
\end{table}

\subsection{Datasets}

\paragraph{SIB-200} \citep{adelani2024sib200} is a topic classification benchmark covering 205~languages with 1,000 examples each, annotated into 7~categories. Examples are parallel across languages (same \texttt{index\_id}), enabling paired cross-lingual comparisons. We use all 25~EU languages plus Ukrainian for fertility measurement, and Ukrainian, Polish, Russian, and Czech for classification.

\paragraph{Ukrainian Wikipedia} We sample 199~articles from the \texttt{wikimedia/wikipedia} dataset (November 2023 dump) as the encyclopedic text register for domain invariance testing (Experiment~1).

\paragraph{ULP} \citep{galeshchuk2024ulp} is an expert-curated benchmark of 347~multiple-choice questions testing Ukrainian grammar and orthography, validated by professional linguists.

\subsection{Evaluation Protocol}

All evaluations use the AWS Bedrock \texttt{invoke\_model} API with temperature~0 for deterministic outputs, maintaining exact consistency with \citet{ovcharov2026tokenizer}. Provider-specific message formatting is preserved: Meta models use the Llama~3/4 prompt template, Amazon Nova uses the \texttt{messages-v1} schema, and remaining models (Qwen, Mistral, NVIDIA) use the standard messages format.

\paragraph{Fertility measurement.} We report the average ratio of API-reported input tokens to whitespace-delimited words. Because SIB-200 texts are short (typically 15--30 words), measuring fertility on individual sentences would be dominated by system prompt overhead. Following \citet{ovcharov2026tokenizer}, we concatenate texts into blocks of approximately 6,000 characters before measurement, yielding blocks of ${\sim}$840 words on average. For EU Acts, we concatenate aligned segments from each language into blocks of similar size. A trivial prompt (``Repeat the first word'') ensures the measurement captures tokenizer behavior on the input text rather than task-specific output.

\paragraph{Classification.} For SIB-200 topic classification, we report accuracy on the test split (204 parallel examples across all languages). Few-shot examples are drawn from the training split: one example per class (7~total for SIB-200). The prompt instructs the model to respond with only the English category name; responses are normalized via substring matching with a Ukrainian-to-English label map to handle models that respond in Ukrainian.

\paragraph{Linguistic competence.} For ULP, we report accuracy on 347 multiple-choice questions in zero-shot mode, and on 344 questions in few-shot mode (3~held out as demonstrations). The prompt presents each question with Ukrainian answer letters (\foreignlanguage{ukrainian}{А, Б, В, Г, Д}) and instructs the model to respond with only the letter.

\section{Experiments and Results}

\subsection{Experiment 1: Cross-Domain Fertility}
\label{sec:exp1}

To test whether the fertility spread observed on legal text is domain-specific or a tokenizer property, we measured fertility on two additional Ukrainian text registers: news (SIB-200, 204 test examples) and encyclopedic text (Ukrainian Wikipedia, 199 articles). All texts were concatenated into ${\sim}$6,000-character blocks matching the protocol of \citet{ovcharov2026tokenizer}. For the six models with publicly available tokenizers, we measured fertility locally; for the remaining four (Mistral Large~3, Nemotron, Nova Pro, Qwen3~235B), we use the API measurements from \citet{ovcharov2026tokenizer}.

Table~\ref{tab:crossdomain_fertility} presents the results across three domains.

\begin{table}[t]
\centering
\caption{Cross-domain tokenizer fertility on Ukrainian text across three registers. Legal from \citet{ovcharov2026tokenizer} (API); News and Wiki measured via local tokenizers. Models sorted by news fertility. \dag~= local tokenizer only (no API baseline).}
\label{tab:crossdomain_fertility}
\begin{tabular}{lR{1.2cm}R{1.2cm}R{1.2cm}}
\toprule
\textbf{Model} & \textbf{Legal} & \textbf{News} & \textbf{Wiki} \\
\midrule
Llama 4 Maverick   & 2.430 & 2.158 & 2.522 \\
Llama 3.3 70B      & 2.650 & 2.317 & 2.674 \\
Gemma 2 27B\dag    & ---   & 2.349 & 2.737 \\
GPT-4o\dag         & ---   & 2.451 & 2.825 \\
Mistral Large 3    & 3.060 & 2.500 & --- \\
Nemotron Super 3   & 3.080 & 2.516 & --- \\
DeepSeek V3\dag    & ---   & 3.068 & 3.403 \\
Nova Pro           & 2.850 & 3.269 & --- \\
Qwen3 32B          & 3.902 & 3.617 & 3.974 \\
\midrule
\textit{Max/Min (6 local)} & & \textit{1.68$\times$} & \textit{1.58$\times$} \\
\bottomrule
\end{tabular}
\end{table}

Three patterns emerge. First, the \textbf{fertility ranking is preserved across all three domains}: the six locally-measured models rank identically on news and encyclopedic text (Spearman $\rho = 1.0$). The max/min ratio is 1.68$\times$ on news and 1.58$\times$ on encyclopedic text, consistent with the 1.61$\times$ on legal text reported by \citet{ovcharov2026tokenizer}. Second, \textbf{encyclopedic text is the most expensive} domain for all six models, reflecting Wikipedia's diverse vocabulary including proper nouns, technical terminology, and transliterated foreign words. Legal text is intermediate, while news is cheapest. Third, the three new models slot predictably into the hierarchy: \textbf{Gemma~2} (256K vocabulary) is nearly as efficient as the Llama family despite a different tokenizer design; \textbf{GPT-4o} (200K vocabulary) falls in the middle tier; \textbf{DeepSeek~V3} (128K vocabulary) clusters with the high-fertility group despite its large vocabulary, suggesting that vocabulary size alone does not determine efficiency on Cyrillic text.

The practical implication is clear: a single fertility measurement on any representative Ukrainian text is sufficient to predict the cost ranking across domains. Practitioners need not re-measure fertility for each new application.

\begin{figure}[t]
\centering
\begin{tikzpicture}
\begin{axis}[
    width=\columnwidth,
    height=6.5cm,
    ybar=2pt,
    bar width=6pt,
    xlabel={},
    ylabel={Fertility (tokens/word)},
    ymin=1.5, ymax=4.5,
    symbolic x coords={Maverick,Llama 3.3,Gemma 2,GPT-4o,DeepSeek,Qwen3},
    xtick=data,
    x tick label style={rotate=30, anchor=east, font=\small},
    legend style={at={(0.02,0.98)}, anchor=north west, font=\small, draw=none, fill=white, fill opacity=0.8},
    legend columns=1,
    ymajorgrids=true,
    grid style={gray!30},
    every axis plot/.append style={fill opacity=0.85},
    nodes near coords style={font=\tiny, rotate=90, anchor=west, xshift=0pt},
]
\addplot[fill=blue!60!black, draw=blue!60!black] coordinates {
    (Maverick, 2.158)
    (Llama 3.3, 2.317)
    (Gemma 2, 2.349)
    (GPT-4o, 2.451)
    (DeepSeek, 3.068)
    (Qwen3, 3.617)
};
\addplot[fill=orange!70!red, draw=orange!70!red] coordinates {
    (Maverick, 2.522)
    (Llama 3.3, 2.674)
    (Gemma 2, 2.737)
    (GPT-4o, 2.825)
    (DeepSeek, 3.403)
    (Qwen3, 3.974)
};
\legend{News (SIB-200), Wiki (encyclopedic)}
\end{axis}
\end{tikzpicture}
\caption{Cross-domain fertility across three Ukrainian text registers. Model rankings are perfectly preserved between news and encyclopedic text ($\rho = 1.0$). Encyclopedic text (Wikipedia) is consistently the most expensive domain due to diverse vocabulary.}
\label{fig:crossdomain}
\end{figure}
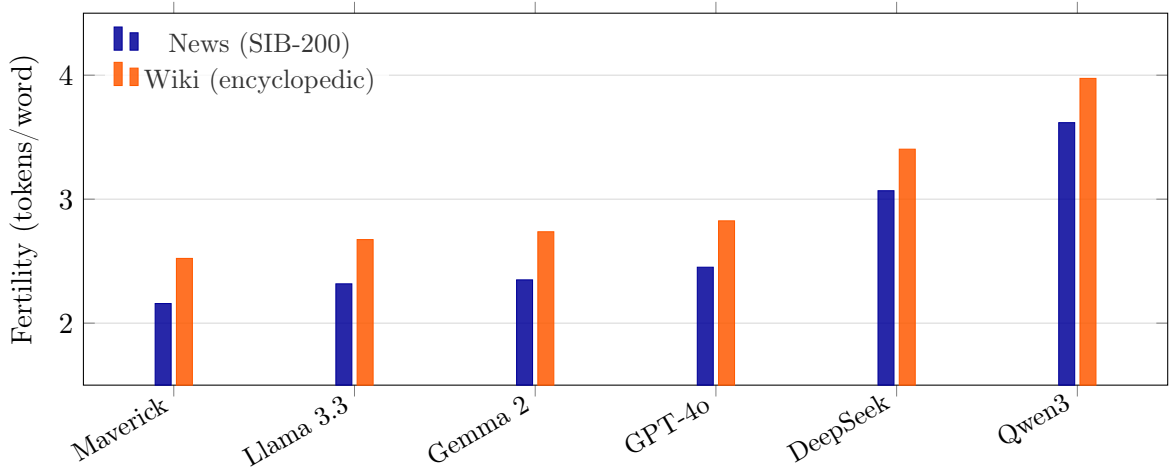

\subsection{Experiment 2: Cross-Lingual Fertility}
\label{sec:exp2}

To contextualize Ukrainian's tokenizer penalty within the European language landscape, we measured fertility for all six locally-available tokenizers across 25~EU languages on SIB-200 parallel text. Because SIB-200 examples are parallel across languages (same \texttt{index\_id}), any fertility difference is attributable solely to the tokenizer's handling of that language, not to content variation.

\begin{figure*}[t]
\centering
\begin{tikzpicture}
\begin{axis}[
    width=0.85\textwidth,
    height=15cm,
    colormap={fertility}{color=(yellow!30) color=(orange) color=(red!70!black) color=(red!30!black)},
    colorbar,
    colorbar style={ylabel=Fertility},
    point meta min=1.0,
    point meta max=5.8,
    xtick={0,...,5},
    xticklabels={Maverick,Llama 3.3,Gemma 2,GPT-4o,DeepSeek,Qwen3},
    x tick label style={rotate=45, anchor=east, font=\small},
    ytick={0,...,24},
    yticklabels={Maltese,Greek,Finnish,Latvian,Lithuanian,Estonian,Hungarian,Ukrainian,Slovak,Bulgarian,Croatian,Slovenian,Czech,Irish,Polish,Romanian,Swedish,Danish,German,Dutch,Italian,Portuguese,French,Spanish,English},
    y tick label style={font=\small},
    xlabel={},
    ylabel={},
    mesh/cols=6,
    nodes near coords={\pgfmathprintnumber[precision=1]{\pgfplotspointmeta}},
    every node near coord/.append style={font=\tiny, black, yshift=-6pt},
]
\addplot[matrix plot*, point meta=explicit] coordinates {
    (0,0) [3.220]
    (1,0) [3.393]
    (2,0) [2.886]
    (3,0) [2.612]
    (4,0) [3.268]
    (5,0) [3.236]
    (0,1) [2.292]
    (1,1) [2.524]
    (2,1) [2.597]
    (3,1) [2.383]
    (4,1) [3.044]
    (5,1) [5.727]
    (0,2) [2.690]
    (1,2) [3.361]
    (2,2) [2.726]
    (3,2) [2.611]
    (4,2) [3.146]
    (5,2) [3.413]
    (0,3) [2.518]
    (1,3) [3.389]
    (2,3) [2.655]
    (3,3) [2.616]
    (4,3) [3.249]
    (5,3) [3.420]
    (0,4) [2.483]
    (1,4) [3.328]
    (2,4) [2.523]
    (3,4) [2.573]
    (4,4) [3.024]
    (5,4) [3.228]
    (0,5) [2.537]
    (1,5) [3.021]
    (2,5) [2.644]
    (3,5) [2.402]
    (4,5) [3.009]
    (5,5) [3.075]
    (0,6) [2.214]
    (1,6) [3.073]
    (2,6) [2.401]
    (3,6) [2.557]
    (4,6) [2.865]
    (5,6) [3.143]
    (0,7) [2.158]
    (1,7) [2.317]
    (2,7) [2.349]
    (3,7) [2.451]
    (4,7) [3.068]
    (5,7) [3.617]
    (0,8) [2.134]
    (1,8) [2.545]
    (2,8) [2.196]
    (3,8) [2.285]
    (4,8) [2.622]
    (5,8) [2.930]
    (0,9) [1.864]
    (1,9) [2.572]
    (2,9) [2.047]
    (3,9) [2.140]
    (4,9) [2.488]
    (5,9) [2.842]
    (0,10) [2.116]
    (1,10) [2.458]
    (2,10) [2.236]
    (3,10) [2.013]
    (4,10) [2.476]
    (5,10) [2.582]
    (0,11) [2.103]
    (1,11) [2.440]
    (2,11) [2.135]
    (3,11) [2.019]
    (4,11) [2.441]
    (5,11) [2.567]
    (0,12) [2.033]
    (1,12) [2.090]
    (2,12) [2.088]
    (3,12) [2.194]
    (4,12) [2.345]
    (5,12) [2.926]
    (0,13) [2.147]
    (1,13) [2.414]
    (2,13) [2.260]
    (3,13) [1.928]
    (4,13) [2.436]
    (5,13) [2.472]
    (0,14) [1.879]
    (1,14) [2.600]
    (2,14) [2.002]
    (3,14) [2.219]
    (4,14) [2.306]
    (5,14) [2.472]
    (0,15) [1.643]
    (1,15) [2.106]
    (2,15) [1.756]
    (3,15) [1.813]
    (4,15) [2.090]
    (5,15) [2.139]
    (0,16) [1.738]
    (1,16) [2.067]
    (2,16) [1.739]
    (3,16) [1.774]
    (4,16) [2.045]
    (5,16) [2.108]
    (0,17) [1.760]
    (1,17) [1.998]
    (2,17) [1.732]
    (3,17) [1.720]
    (4,17) [2.000]
    (5,17) [2.038]
    (0,18) [1.578]
    (1,18) [1.967]
    (2,18) [1.549]
    (3,18) [1.587]
    (4,18) [1.873]
    (5,18) [1.980]
    (0,19) [1.551]
    (1,19) [1.864]
    (2,19) [1.560]
    (3,19) [1.438]
    (4,19) [1.744]
    (5,19) [1.892]
    (0,20) [1.491]
    (1,20) [1.835]
    (2,20) [1.501]
    (3,20) [1.625]
    (4,20) [1.719]
    (5,20) [1.861]
    (0,21) [1.379]
    (1,21) [1.692]
    (2,21) [1.403]
    (3,21) [1.375]
    (4,21) [1.619]
    (5,21) [1.699]
    (0,22) [1.402]
    (1,22) [1.655]
    (2,22) [1.402]
    (3,22) [1.396]
    (4,22) [1.594]
    (5,22) [1.665]
    (0,23) [1.329]
    (1,23) [1.597]
    (2,23) [1.307]
    (3,23) [1.339]
    (4,23) [1.547]
    (5,23) [1.611]
    (0,24) [1.217]
    (1,24) [1.223]
    (2,24) [1.232]
    (3,24) [1.213]
    (4,24) [1.218]
    (5,24) [1.251]
};
\end{axis}
\end{tikzpicture}
\caption{Tokenizer fertility across 25~languages and 6~models on SIB-200 parallel text. Languages sorted by mean fertility. Ukrainian (2.66 mean) is more expensive than all other Slavic languages. Qwen3's Greek penalty (5.73) is an outlier; other models handle Greek at 2.3--3.0.}
\label{fig:eu_heatmap}
\end{figure*}
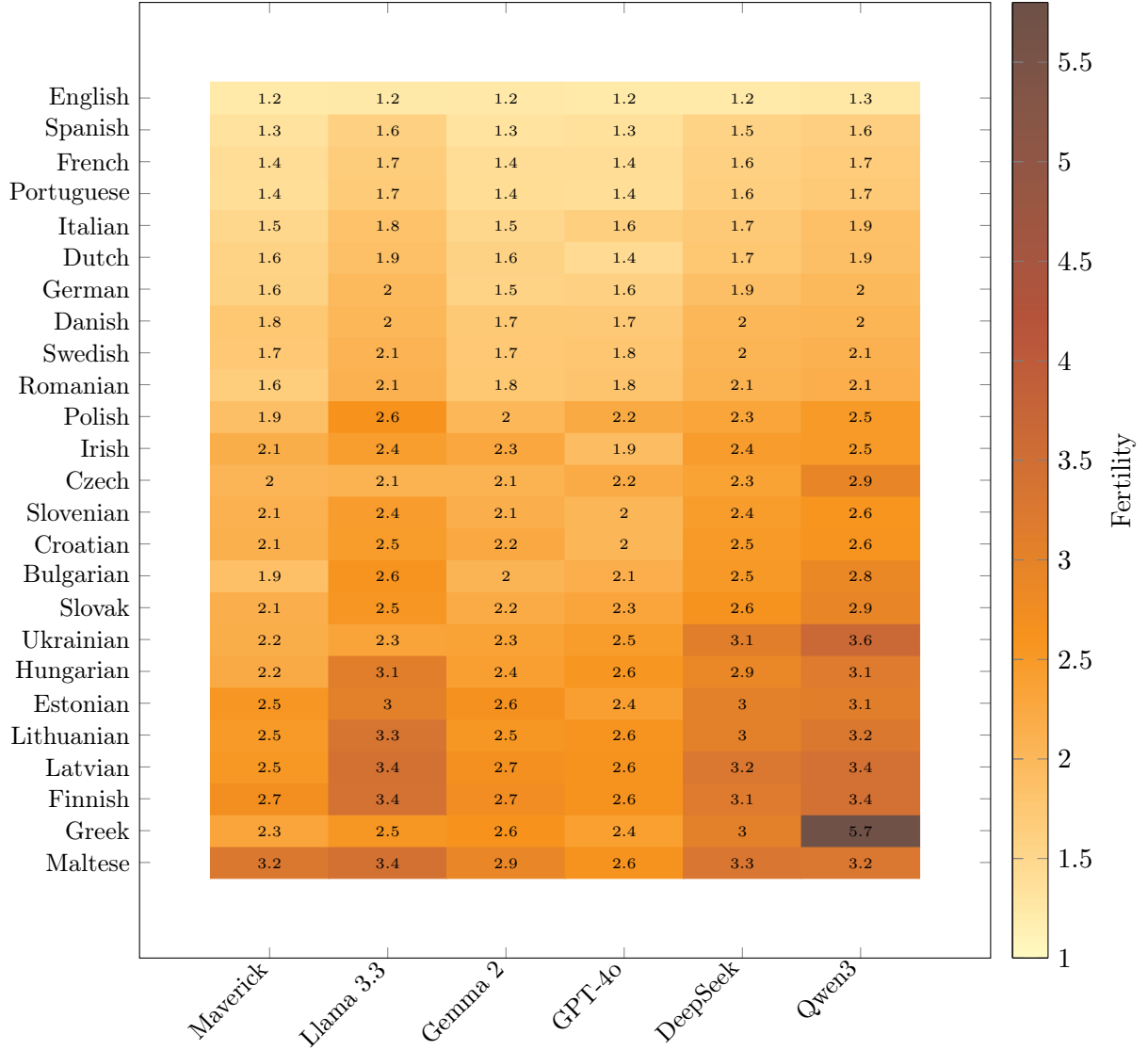

Table~\ref{tab:eu_fertility} presents mean fertility by language family. Figure~\ref{fig:eu_heatmap} shows the full 25-language $\times$ 6-model heatmap.

\begin{table}[t]
\centering
\caption{Mean tokenizer fertility across 6~models on SIB-200 parallel text, by language family. Languages sorted by mean fertility within each family. Min/Max columns show the range across models.}
\label{tab:eu_fertility}
\begin{tabular}{llR{1.2cm}R{1.0cm}R{1.0cm}}
\toprule
\textbf{Family} & \textbf{Language} & \textbf{Mean} & \textbf{Min} & \textbf{Max} \\
\midrule
English & English & 1.23 & 1.21 & 1.25 \\
\midrule
\multirow{4}{*}{Romance} & Spanish & 1.46 & 1.31 & 1.61 \\
                          & French & 1.52 & 1.40 & 1.67 \\
                          & Portuguese & 1.53 & 1.38 & 1.70 \\
                          & Italian & 1.67 & 1.49 & 1.86 \\
                          & Romanian & 1.92 & 1.64 & 2.14 \\
\midrule
\multirow{4}{*}{Germanic} & Dutch & 1.67 & 1.44 & 1.89 \\
                           & German & 1.76 & 1.55 & 1.98 \\
                           & Danish & 1.87 & 1.72 & 2.04 \\
                           & Swedish & 1.91 & 1.74 & 2.11 \\
\midrule
Celtic & Irish & 2.28 & 1.93 & 2.47 \\
\midrule
\multirow{6}{*}{Slavic} & Polish & 2.25 & 1.88 & 2.60 \\
                         & Czech & 2.28 & 2.03 & 2.93 \\
                         & Slovenian & 2.28 & 2.02 & 2.57 \\
                         & Croatian & 2.31 & 2.01 & 2.58 \\
                         & Bulgarian & 2.33 & 1.86 & 2.84 \\
                         & Slovak & 2.45 & 2.13 & 2.93 \\
\midrule
\multirow{3}{*}{Uralic} & Estonian & 2.78 & 2.40 & 3.08 \\
                         & Hungarian & 2.71 & 2.21 & 3.14 \\
                         & Finnish & 2.99 & 2.61 & 3.41 \\
\midrule
\multirow{2}{*}{Baltic} & Lithuanian & 2.86 & 2.48 & 3.33 \\
                         & Latvian & 2.97 & 2.52 & 3.42 \\
\midrule
\textbf{Target} & \textbf{Ukrainian} & \textbf{2.66} & \textbf{2.16} & \textbf{3.62} \\
\midrule
Hellenic & Greek & 3.09 & 2.29 & 5.73 \\
Semitic & Maltese & 3.10 & 2.61 & 3.39 \\
\bottomrule
\end{tabular}
\end{table}

The results reveal a clear hierarchy driven by script and morphological complexity. Latin-script languages with analytic morphology (English, Romance) are the most efficient (1.2--1.7 mean tokens/word). Germanic languages cluster at 1.7--1.9. Slavic languages span 2.2--2.5. Agglutinative and morphologically complex languages (Uralic, Baltic) reach 2.7--3.0. Greek and Maltese are the most expensive at ${\sim}$3.1.

Ukrainian (2.66 mean) is more expensive to tokenize than any other Slavic language in the dataset, despite similar morphological complexity to Polish (2.25) or Czech (2.28). This gap -- approximately 15--18\% higher than cognate Slavic languages -- suggests that Ukrainian's disadvantage is not purely morphological but also reflects \textbf{underrepresentation in pre-training data}. Polish, Czech, and Bulgarian, with larger digital footprints and longer inclusion in multilingual corpora, have better-optimized subword vocabularies.

The Min/Max columns in Table~\ref{tab:eu_fertility} reveal that model variation is substantial: Greek ranges from 2.29 (Maverick) to 5.73 (Qwen3), a 2.5$\times$ spread for the \emph{same language}. Ukrainian shows a similar pattern: 2.16 (Maverick) to 3.62 (Qwen3), confirming that the Ukrainian penalty depends strongly on tokenizer design. Maverick and Gemma~2 handle Cyrillic efficiently; Qwen3 and DeepSeek do not.

Figure~\ref{fig:eu_heatmap} shows the full heatmap. The hierarchy is consistent across models: English is always cheapest; Greek, Maltese, and Ukrainian cluster at the expensive end. Model rankings are stable across all languages: Maverick and GPT-4o are consistently the most efficient, while Qwen3 is the least efficient.

\subsection{Experiment 2b: Subword Decomposition Analysis}
\label{sec:exp2b}

To understand \emph{why} fertility varies across tokenizers on the same Ukrainian text, we examined how each tokenizer splits 12~representative legal terms. Table~\ref{tab:subword} shows the decomposition for six models on four high-frequency words.

\begin{table*}[t]
\centering
\caption{Subword decomposition of Ukrainian legal terms. Pipe ($|$) separates subword tokens. $n$ = token count. Models ordered by vocabulary size. Transliterated for readability.}
\label{tab:subword}
\begin{tabular}{lp{4.5cm}cp{4.5cm}c}
\toprule
\textbf{Model (vocab)} & \foreignlanguage{ukrainian}{\textbf{відповідальність}} \textit{(responsibility)} & $n$ & \foreignlanguage{ukrainian}{\textbf{підприємництва}} \textit{(entrepreneurship)} & $n$ \\
\midrule
Gemma 2 (256K)  & vid$|$povi$|$dal$|$nist & 4 & pid$|$pryyem$|$nyctva & 3 \\
Maverick (200K) & vid$|$pov$|$i$|$dal$|$nist & 5 & pid$|$pryyem$|$nyctva & 3 \\
GPT-4o (200K)   & vid$|$pov$|$id$|$al$|$nist & 5 & p$|$id$|$pryyem$|$nyc$|$tva & 5 \\
Llama 3.3 (128K) & vid$|$povid$|$al$|$nist & 4 & pid$|$pryyem$|$nyctva & 3 \\
DeepSeek (128K) & v$|$id$|$pov$|$i$|$da$|$l$|$ni$|$st & 8 & p$|$id$|$pry$|$ye$|$m$|$nyc$|$t$|$va & 8 \\
Qwen3 (151K)    & v$|$i$|$d$|$pov$|$i$|$d$|$aln$|$i$|$st & 9 & p$|$i$|$d$|$pry$|$ye$|$m$|$nyc$|$t$|$va & 9 \\
\bottomrule
\end{tabular}
\end{table*}

Two patterns are clear. First, \textbf{low-fertility tokenizers preserve morpheme boundaries}: Gemma~2 and Llama~3.3 split \foreignlanguage{ukrainian}{``відповідальність''} into recognizable morphemes, while Qwen3 and DeepSeek fragment it into near-character-level pieces that cross morphological boundaries.

Second, \textbf{vocabulary size alone does not predict efficiency}: DeepSeek~V3 (128K vocab) and Qwen3 (151K vocab) both fragment Ukrainian words into 8--9 tokens, while Llama~3.3 (128K vocab) achieves 3--4 tokens. The proportion of Cyrillic-specific merge operations matters more than total vocabulary size. Models trained predominantly on English and Chinese text (Qwen, DeepSeek) allocate fewer vocabulary entries to Ukrainian morphemes despite large total vocabularies.

Across all 12~analyzed words, the mean token count is: Maverick 3.5, Llama~3.3 3.6, Gemma~2 3.7, GPT-4o 4.2, DeepSeek 5.4, Qwen3 6.6. The 1.9$\times$ ratio between the most and least efficient tokenizers on individual words is consistent with the corpus-level fertility measurements in Experiments~1 and 2.

\subsection{Experiment 3: Cross-Lingual Few-Shot Classification}
\label{sec:exp3}

To test whether the few-shot effect is language-specific or model-intrinsic, we evaluated all seven models on SIB-200 topic classification across four Slavic languages using parallel examples (same \texttt{index\_id}). We also include the legal text few-shot delta from \citet{ovcharov2026tokenizer} for cross-domain comparison.

\begin{table*}[t]
\centering
\caption{Few-shot delta ($\Delta_{\text{FS}}$ = FS $-$ ZS accuracy, in pp) on SIB-200 topic classification across four Slavic languages, plus Ukrainian legal text from \citet{ovcharov2026tokenizer}. Models sorted by SIB-200 average. Bold = consistent direction across all 4 news languages.}
\label{tab:slavic}
\begin{tabular}{lR{1.0cm}R{1.0cm}R{1.0cm}R{1.0cm}R{1.0cm}R{1.4cm}}
\toprule
& \multicolumn{5}{c}{\textbf{SIB-200 (news)}} & \textbf{Legal} \\
\cmidrule(lr){2-6}
\textbf{Model} & \textbf{UK} & \textbf{PL} & \textbf{RU} & \textbf{CZ} & \textbf{Avg} & \textbf{UK} \\
\midrule
\textbf{Nemotron Super 3}  & $+$8.3  & $+$15.7 & $+$8.8  & $+$9.3  & \textbf{$+$10.5} & $-$12.8 \\
Qwen3 32B         & $+$3.9  & $+$5.9  & $+$7.4  & $+$7.8  & \textbf{$+$6.2} & $-$6.6 \\
Nova Pro          & $+$4.4  & $+$5.1  & $+$3.0  & $+$1.5  & \textbf{$+$3.5} & $-$2.6 \\
Qwen3 235B        & $+$4.9  & $+$3.4  & $+$1.5  & $+$2.9  & \textbf{$+$3.2} & $-$26.0 \\
Llama 3.3 70B     & $+$3.4  & $+$3.0  & $-$0.5  & $+$5.0  & $+$2.7 & $+$0.4 \\
Mistral Large 3   & $-$4.4  & $-$1.0  & $+$2.9  & $+$1.0  & $-$0.4 & $-$3.3 \\
\textbf{Llama 4 Maverick}  & $-$7.8  & $-$6.3  & $-$15.7 & $-$3.8  & \textbf{$-$8.4} & $-$6.2 \\
\bottomrule
\end{tabular}
\end{table*}

With 204 test examples and 7~classes, the 95\% Clopper--Pearson confidence interval on a single accuracy is $\pm$5--7~pp. To assess whether few-shot deltas exceed this noise floor, we compute exact McNemar's tests on paired zero-shot vs.\ few-shot predictions for each model$\times$language cell. Of the 28~cells (7~models $\times$ 4~languages), 12~are significant at $p < 0.05$ after Holm--Bonferroni correction: all four Maverick cells (degradation) and all four Nemotron cells (improvement), plus Qwen3~32B on Russian and Czech, and Llama~3.3 on Czech. The remaining deltas, while directionally consistent, do not individually reach significance at this sample size.

The results answer the central question of this paper. The few-shot effect pattern is \textbf{identical across all four Slavic languages}:

\begin{itemize}[leftmargin=*]
    \item Four models (Nemotron, Qwen3~32B, Qwen3~235B, Nova~Pro) \textbf{improve on all four languages}---average deltas range from $+$3.2 to $+$10.5~pp. Nemotron's improvement is significant on all four ($p < 0.01$).
    \item One model (Llama~4 Maverick) \textbf{degrades on all four languages}---average $-$8.4~pp, with Russian showing the most severe degradation ($-$15.7~pp). All four Maverick cells are significant ($p < 0.001$).
    \item Mistral Large~3 and Llama~3.3 show mixed effects, with language-dependent direction; individual deltas are not significant.
\end{itemize}

By language, 5--6 of 7 models improve with few-shot examples across all four languages (Czech: 6/7, others: 5/7). The consistency across languages with different scripts (Cyrillic for Ukrainian and Russian vs.\ Latin for Polish and Czech), different tokenizer fertility levels, and different pre-training data volumes rules out both the ``Ukrainian-specific'' and ``morphological complexity'' hypotheses.

The few-shot effect is \textbf{model-intrinsic}: Maverick's attention mechanism consistently anchors on surface patterns in demonstrations regardless of language, while Nemotron's hybrid architecture consistently leverages demonstrations for task specification. This is a property of model design, not of the input language.

\paragraph{Error analysis: Maverick degradation.} To understand \emph{how} Maverick degrades, we examined its Ukrainian SIB-200 predictions. In zero-shot mode, errors are distributed across classes roughly proportionally to class frequency. In few-shot mode, Maverick shifts toward two dominant classes---\emph{science\_and\_technology} and \emph{politics}---at the expense of minority classes: \emph{geography} recall drops from 76.5\% to 41.2\%, and \emph{entertainment} from 82.1\% to 64.3\%. This pattern is consistent with \emph{demonstration anchoring}: the model over-weights surface patterns from the provided examples (which include one geography and one entertainment example) rather than using them for task specification. The same class-collapse pattern appears on Russian ($-$15.7~pp), where Maverick's geography recall drops from 70.6\% to 23.5\%. This is consistent with \emph{demonstration anchoring}: hidden-state representations shift toward the demonstration content rather than the query \citep{ovcharov2026tokenizer}.

\begin{figure*}[t]
\centering
\begin{tikzpicture}[x=1pt,y=1pt]
\definecolor{fillColor}{RGB}{255,255,255}
\path[use as bounding box,fill=fillColor,fill opacity=0.00] (0,0) rectangle (397.48,216.81);
\begin{scope}
\path[clip] (  0.00,  0.00) rectangle (397.48,216.81);
\definecolor{fillColor}{RGB}{255,255,255}

\path[fill=fillColor] (  0.00,  0.00) rectangle (397.48,216.81);
\end{scope}
\begin{scope}
\path[clip] ( 44.19, 19.96) rectangle (128.01,201.33);
\definecolor{drawColor}{gray}{0.92}

\path[draw=drawColor,line width= 0.4pt,line join=round] ( 61.83, 19.96) --
	( 61.83,201.33);

\path[draw=drawColor,line width= 0.4pt,line join=round] ( 86.10, 19.96) --
	( 86.10,201.33);

\path[draw=drawColor,line width= 0.4pt,line join=round] (110.37, 19.96) --
	(110.37,201.33);
\definecolor{fillColor}{RGB}{192,57,43}

\path[fill=fillColor] ( 67.17, 26.26) rectangle ( 86.10, 43.89);
\definecolor{fillColor}{RGB}{39,174,96}

\path[fill=fillColor] ( 86.10, 76.64) rectangle ( 94.35, 94.28);
\definecolor{fillColor}{RGB}{192,57,43}

\path[fill=fillColor] ( 75.42, 51.45) rectangle ( 86.10, 69.08);
\definecolor{fillColor}{RGB}{39,174,96}

\path[fill=fillColor] ( 86.10,177.40) rectangle (106.24,195.04);

\path[fill=fillColor] ( 86.10,127.02) rectangle ( 96.78,144.66);

\path[fill=fillColor] ( 86.10,101.83) rectangle ( 97.99,119.47);

\path[fill=fillColor] ( 86.10,152.21) rectangle ( 95.56,169.85);
\definecolor{drawColor}{gray}{0.30}

\path[draw=drawColor,line width= 0.3pt,line join=round] ( 86.10, 19.96) -- ( 86.10,201.33);
\end{scope}
\begin{scope}
\path[clip] (132.01, 19.96) rectangle (215.84,201.33);
\definecolor{drawColor}{gray}{0.92}

\path[draw=drawColor,line width= 0.4pt,line join=round] (149.66, 19.96) --
	(149.66,201.33);

\path[draw=drawColor,line width= 0.4pt,line join=round] (173.92, 19.96) --
	(173.92,201.33);

\path[draw=drawColor,line width= 0.4pt,line join=round] (198.19, 19.96) --
	(198.19,201.33);
\definecolor{fillColor}{RGB}{192,57,43}

\path[fill=fillColor] (158.63, 26.26) rectangle (173.92, 43.89);
\definecolor{fillColor}{RGB}{39,174,96}

\path[fill=fillColor] (173.92, 76.64) rectangle (181.20, 94.28);
\definecolor{fillColor}{RGB}{192,57,43}

\path[fill=fillColor] (171.50, 51.45) rectangle (173.92, 69.08);
\definecolor{fillColor}{RGB}{39,174,96}

\path[fill=fillColor] (173.92,177.40) rectangle (212.03,195.04);

\path[fill=fillColor] (173.92,127.02) rectangle (186.30,144.66);

\path[fill=fillColor] (173.92,101.83) rectangle (182.18,119.47);

\path[fill=fillColor] (173.92,152.21) rectangle (188.24,169.85);
\definecolor{drawColor}{gray}{0.30}

\path[draw=drawColor,line width= 0.3pt,line join=round] (173.92, 19.96) -- (173.92,201.33);
\end{scope}
\begin{scope}
\path[clip] (219.84, 19.96) rectangle (303.66,201.33);
\definecolor{drawColor}{gray}{0.92}

\path[draw=drawColor,line width= 0.4pt,line join=round] (237.48, 19.96) --
	(237.48,201.33);

\path[draw=drawColor,line width= 0.4pt,line join=round] (261.75, 19.96) --
	(261.75,201.33);

\path[draw=drawColor,line width= 0.4pt,line join=round] (286.02, 19.96) --
	(286.02,201.33);
\definecolor{fillColor}{RGB}{192,57,43}

\path[fill=fillColor] (223.65, 26.26) rectangle (261.75, 43.89);

\path[fill=fillColor] (260.54, 76.64) rectangle (261.75, 94.28);
\definecolor{fillColor}{RGB}{39,174,96}

\path[fill=fillColor] (261.75, 51.45) rectangle (268.79, 69.08);

\path[fill=fillColor] (261.75,177.40) rectangle (283.11,195.04);

\path[fill=fillColor] (261.75,127.02) rectangle (269.03,144.66);

\path[fill=fillColor] (261.75,101.83) rectangle (265.39,119.47);

\path[fill=fillColor] (261.75,152.21) rectangle (279.71,169.85);
\definecolor{drawColor}{gray}{0.30}

\path[draw=drawColor,line width= 0.3pt,line join=round] (261.75, 19.96) -- (261.75,201.33);
\end{scope}
\begin{scope}
\path[clip] (307.66, 19.96) rectangle (391.48,201.33);
\definecolor{drawColor}{gray}{0.92}

\path[draw=drawColor,line width= 0.4pt,line join=round] (325.30, 19.96) --
	(325.30,201.33);

\path[draw=drawColor,line width= 0.4pt,line join=round] (349.57, 19.96) --
	(349.57,201.33);

\path[draw=drawColor,line width= 0.4pt,line join=round] (373.84, 19.96) --
	(373.84,201.33);
\definecolor{fillColor}{RGB}{192,57,43}

\path[fill=fillColor] (340.35, 26.26) rectangle (349.57, 43.89);
\definecolor{fillColor}{RGB}{39,174,96}

\path[fill=fillColor] (349.57, 76.64) rectangle (361.71, 94.28);

\path[fill=fillColor] (349.57, 51.45) rectangle (352.00, 69.08);

\path[fill=fillColor] (349.57,177.40) rectangle (372.14,195.04);

\path[fill=fillColor] (349.57,127.02) rectangle (353.21,144.66);

\path[fill=fillColor] (349.57,101.83) rectangle (356.61,119.47);

\path[fill=fillColor] (349.57,152.21) rectangle (368.50,169.85);
\definecolor{drawColor}{gray}{0.30}

\path[draw=drawColor,line width= 0.3pt,line join=round] (349.57, 19.96) -- (349.57,201.33);
\end{scope}
\begin{scope}
\path[clip] ( 44.19,201.33) rectangle (128.01,214.81);
\definecolor{drawColor}{gray}{0.10}

\node[text=drawColor,anchor=base,inner sep=0pt, outer sep=0pt, scale=  0.80] at ( 86.10,205.31) {\bfseries Ukrainian};
\end{scope}
\begin{scope}
\path[clip] (132.01,201.33) rectangle (215.84,214.81);
\definecolor{drawColor}{gray}{0.10}

\node[text=drawColor,anchor=base,inner sep=0pt, outer sep=0pt, scale=  0.80] at (173.92,205.31) {\bfseries Polish};
\end{scope}
\begin{scope}
\path[clip] (219.84,201.33) rectangle (303.66,214.81);
\definecolor{drawColor}{gray}{0.10}

\node[text=drawColor,anchor=base,inner sep=0pt, outer sep=0pt, scale=  0.80] at (261.75,205.31) {\bfseries Russian};
\end{scope}
\begin{scope}
\path[clip] (307.66,201.33) rectangle (391.48,214.81);
\definecolor{drawColor}{gray}{0.10}

\node[text=drawColor,anchor=base,inner sep=0pt, outer sep=0pt, scale=  0.80] at (349.57,205.31) {\bfseries Czech};
\end{scope}
\begin{scope}
\path[clip] (  0.00,  0.00) rectangle (397.48,216.81);
\definecolor{drawColor}{gray}{0.30}

\node[text=drawColor,anchor=base,inner sep=0pt, outer sep=0pt, scale=  0.60] at ( 61.83, 12.23) {-10};

\node[text=drawColor,anchor=base,inner sep=0pt, outer sep=0pt, scale=  0.60] at ( 86.10, 12.23) {0};

\node[text=drawColor,anchor=base,inner sep=0pt, outer sep=0pt, scale=  0.60] at (110.37, 12.23) {10};
\end{scope}
\begin{scope}
\path[clip] (  0.00,  0.00) rectangle (397.48,216.81);
\definecolor{drawColor}{gray}{0.30}

\node[text=drawColor,anchor=base,inner sep=0pt, outer sep=0pt, scale=  0.60] at (149.66, 12.23) {-10};

\node[text=drawColor,anchor=base,inner sep=0pt, outer sep=0pt, scale=  0.60] at (173.92, 12.23) {0};

\node[text=drawColor,anchor=base,inner sep=0pt, outer sep=0pt, scale=  0.60] at (198.19, 12.23) {10};
\end{scope}
\begin{scope}
\path[clip] (  0.00,  0.00) rectangle (397.48,216.81);
\definecolor{drawColor}{gray}{0.30}

\node[text=drawColor,anchor=base,inner sep=0pt, outer sep=0pt, scale=  0.60] at (237.48, 12.23) {-10};

\node[text=drawColor,anchor=base,inner sep=0pt, outer sep=0pt, scale=  0.60] at (261.75, 12.23) {0};

\node[text=drawColor,anchor=base,inner sep=0pt, outer sep=0pt, scale=  0.60] at (286.02, 12.23) {10};
\end{scope}
\begin{scope}
\path[clip] (  0.00,  0.00) rectangle (397.48,216.81);
\definecolor{drawColor}{gray}{0.30}

\node[text=drawColor,anchor=base,inner sep=0pt, outer sep=0pt, scale=  0.60] at (325.30, 12.23) {-10};

\node[text=drawColor,anchor=base,inner sep=0pt, outer sep=0pt, scale=  0.60] at (349.57, 12.23) {0};

\node[text=drawColor,anchor=base,inner sep=0pt, outer sep=0pt, scale=  0.60] at (373.84, 12.23) {10};
\end{scope}
\begin{scope}
\path[clip] (  0.00,  0.00) rectangle (397.48,216.81);
\definecolor{drawColor}{gray}{0.30}

\node[text=drawColor,anchor=base east,inner sep=0pt, outer sep=0pt, scale=  0.70] at ( 40.59, 32.67) {Maverick};

\node[text=drawColor,anchor=base east,inner sep=0pt, outer sep=0pt, scale=  0.70] at ( 40.59, 57.86) {Mistral};

\node[text=drawColor,anchor=base east,inner sep=0pt, outer sep=0pt, scale=  0.70] at ( 40.59, 83.05) {Llama 3.3};

\node[text=drawColor,anchor=base east,inner sep=0pt, outer sep=0pt, scale=  0.70] at ( 40.59,108.24) {Qwen3 235B};

\node[text=drawColor,anchor=base east,inner sep=0pt, outer sep=0pt, scale=  0.70] at ( 40.59,133.43) {Nova Pro};

\node[text=drawColor,anchor=base east,inner sep=0pt, outer sep=0pt, scale=  0.70] at ( 40.59,158.62) {Qwen3 32B};

\node[text=drawColor,anchor=base east,inner sep=0pt, outer sep=0pt, scale=  0.70] at ( 40.59,183.81) {Nemotron};
\end{scope}
\begin{scope}
\path[clip] (  0.00,  0.00) rectangle (397.48,216.81);
\definecolor{drawColor}{RGB}{0,0,0}

\node[text=drawColor,anchor=base,inner sep=0pt, outer sep=0pt, scale=  0.80] at (217.84,  3.56) {$\Delta_{\mathrm{FS}}$ (pp)};
\end{scope}
\end{tikzpicture}
\caption{Few-shot delta ($\Delta_{\text{FS}}$, pp) across four Slavic languages on SIB-200. Red bars = degradation, green = improvement. Maverick degrades on all four languages; Nemotron improves on all four. The pattern is model-intrinsic, not language-dependent.}
\label{fig:fewshot_delta}
\end{figure*}
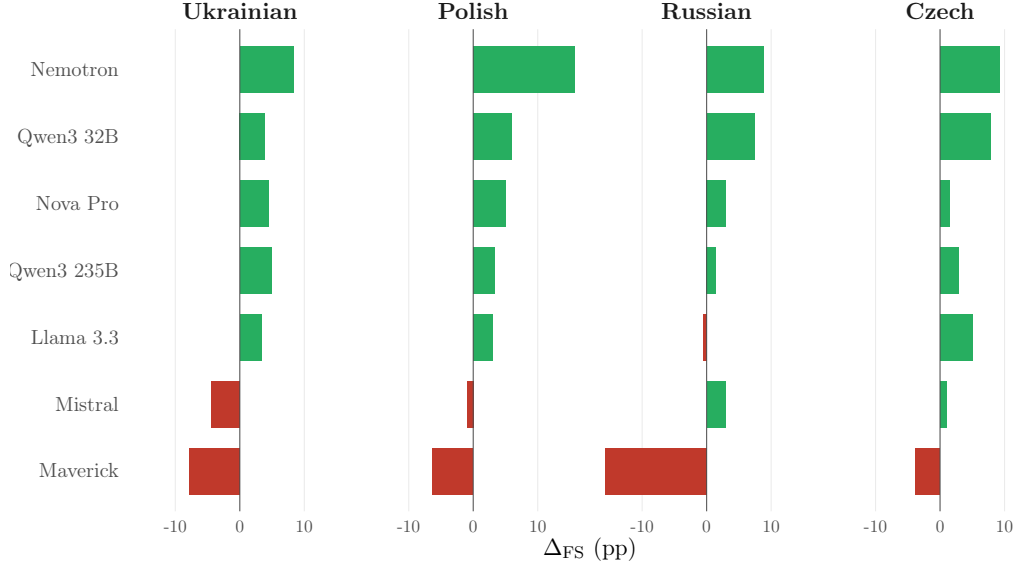

\subsection{Experiment 4: Linguistic Competence}
\label{sec:exp4}

To test whether tokenizer fertility predicts broader linguistic capability, we evaluated all seven models on the ULP benchmark---347 expert-curated multiple-choice questions testing Ukrainian grammar and orthography.

\begin{table}[t]
\centering
\caption{ULP Ukrainian Language Proficiency results. Fertility from \citet{ovcharov2026tokenizer}. $\Delta_{\text{FS}}$ = few-shot minus zero-shot accuracy.}
\label{tab:ulp}
\begin{tabular}{lR{1.2cm}R{1.2cm}R{1.2cm}R{1.4cm}}
\toprule
\textbf{Model} & \textbf{Fert.} & \textbf{ZS} & \textbf{FS} & \textbf{$\Delta_{\text{FS}}$} \\
\midrule
Llama 4 Maverick   & 2.43 & 57.3\% & 53.2\% & $-$4.2 \\
Llama 3.3 70B      & 2.65 & 33.4\% & 36.9\% & $+$3.5 \\
Nova Pro           & 2.85 & 42.9\% & 41.6\% & $-$1.4 \\
Mistral Large 3    & 3.06 & 51.0\% & 49.4\% & $-$1.6 \\
Nemotron Super 3   & 3.08 & 27.7\% & 36.3\% & $+$8.7 \\
Qwen3 235B         & 3.89 & 42.1\% & 44.2\% & $+$2.1 \\
Qwen3 32B          & 3.90 & 34.6\% & 31.7\% & $-$2.9 \\
\bottomrule
\end{tabular}
\end{table}

Llama~4 Maverick---the model with the most efficient tokenizer (fertility 2.43)---achieves the highest ULP accuracy (57.3\%), 6~percentage points above the next-best model (Mistral Large~3, 51.0\%). However, the overall correlation between fertility and zero-shot accuracy is weak: Spearman $\rho = -0.43$ ($p = 0.34$), Pearson $r = -0.35$ ($p = 0.44$). With only seven data points, statistical power is limited, but the trend direction is consistent with the hypothesis that better tokenization facilitates linguistic competence.

Nemotron Super~3 is a notable outlier: despite moderate fertility (3.08), it achieves only 27.7\% zero-shot accuracy---the lowest among all models, and barely above the 20--25\% random baseline for 4--5 choice questions. This model's hybrid Mamba-Transformer architecture may be optimized for long-document reasoning rather than fine-grained grammatical knowledge.

The few-shot effect on ULP is mixed: four models degrade, three improve. The largest improvement is Nemotron Super~3 ($+$8.7~pp), partially compensating for its low zero-shot baseline. The largest degradation is Maverick ($-$4.2~pp), consistent with its pattern of few-shot sensitivity observed on SIB-200.

\section{Discussion}

\subsection{Fertility Is a Tokenizer Property}

Experiment~1 demonstrates that tokenizer fertility rankings are invariant across domains. The 1.68$\times$ spread between the most and least efficient tokenizers on news text is consistent with the 1.58$\times$ spread on encyclopedic text and the 1.61$\times$ on legal text \citep{ovcharov2026tokenizer}. Model rankings are perfectly preserved ($\rho = 1.0$): Qwen3 is the least efficient (${\sim}$3.6 tokens/word on news), Maverick the most efficient (${\sim}$2.2), with Gemma~2, GPT-4o, and DeepSeek in between.

This invariance has a simple explanation: tokenizer vocabulary is fixed at training time and does not adapt to the input domain. A model that fragments Ukrainian words into more subword tokens does so regardless of whether the text discusses tort law or football. The absolute fertility level is lower on news text (6 of 7 models show a negative $\Delta$), reflecting the higher density of domain-specific terminology in legal text, but the \emph{relative ranking} is determined by vocabulary design.

Experiment~2 extends this finding across 25~EU languages: mean fertility varies by 2.5$\times$ between English (1.23) and Greek/Maltese (${\sim}$3.1), with substantial model variation (Qwen3 reaches 5.73 on Greek). Within the Slavic family, Ukrainian (2.66) is 15--18\% more expensive than Polish (2.25) or Czech (2.28), suggesting that Ukrainian's penalty combines morphological complexity with underrepresentation in pre-training data. This has direct cost implications: processing the same content in Ukrainian costs 2.2$\times$ more than in English on average, purely due to tokenizer design.

\subsection{Few-Shot Effects Are Model-Intrinsic, Confirmed Cross-Lingually}

Experiment~3 confirms on API models what \citet{ovcharov2026tokenizer} observed on Ukrainian legal text: the few-shot effect is task-dependent, with five of seven models improving on SIB-200 news while five of seven degraded on legal text. Two models -- Maverick ($-$7.8~pp) and Mistral ($-$4.4~pp) -- degrade on both tasks.

The cross-lingual dimension (Experiment~3) adds a finding that single-language studies cannot provide: the effect is not just task-dependent but \textbf{model-intrinsic across languages}. Maverick degrades on all four Slavic languages ($-$8.4~pp average); Nemotron improves on all four ($+$10.5~pp). This cross-lingual consistency forms a complete picture: the few-shot effect is determined by how a model processes demonstration content, not by the input language.

\subsection{The Ukrainian Penalty: Morphology Plus Underrepresentation}

Experiment~2 reveals that Ukrainian's tokenizer tax is not explained by morphological complexity alone. Within the Slavic family, Polish (2.25) and Czech (2.28) -- languages with comparable inflectional systems -- are 15--18\% cheaper to tokenize than Ukrainian (2.66). This gap likely reflects differences in digital presence and pre-training data volume.

Publicly available corpus statistics confirm this hypothesis. Table~\ref{tab:pretraining} shows the volume of Ukrainian vs.\ other Slavic languages in three major pre-training corpora. Despite comparable speaker populations (${\sim}$38--40M for Ukrainian and Polish), Ukrainian consistently has 2--6$\times$ less training data: 3.1$\times$ fewer tokens than Polish in CulturaX \citep{nguyen2024culturax}, 2.4$\times$ less text in mC4, and 5.7$\times$ fewer words in OSCAR~23.01 \citep{abadji2022towards}. Czech, with only 10.5M speakers, has 1.2--3.0$\times$ more data than Ukrainian in most corpora. This data deficit directly explains the tokenizer penalty: subword vocabularies are optimized on training data, and languages with less data receive fewer dedicated merge operations, resulting in more fragmented tokenization.

\begin{table}[t]
\centering
\caption{Ukrainian vs.\ other Slavic languages in major pre-training corpora. Ratio = language volume / Ukrainian volume.}
\label{tab:pretraining}
\begin{tabular}{lR{1.4cm}R{1.0cm}R{1.4cm}R{1.0cm}}
\toprule
\textbf{Corpus} & \textbf{UK} & \textbf{PL} & \textbf{CZ} & \textbf{RU} \\
\midrule
mC4 (GB)       & 196   & 473 (2.4$\times$)  & 235 (1.2$\times$)  & 3{,}615 (18$\times$) \\
OSCAR 23.01 (B words) & 3.2  & 18.1 (5.7$\times$) & 9.7 (3.0$\times$)  & 78.0 (24$\times$) \\
CulturaX (B tokens)   & 38.2 & 117.3 (3.1$\times$) & 56.9 (1.5$\times$) & 737.2 (19$\times$) \\
\bottomrule
\end{tabular}
\end{table}

The practical consequence is concrete: processing the same content in Ukrainian costs 2.2$\times$ more than in English on average, purely due to tokenizer design. For production systems serving Ukrainian users, this tax compounds across every API call.

\subsection{Fertility and Linguistic Competence}

Experiment~4 is exploratory: with $n = 7$ models, it cannot establish a robust fertility--competence relationship. The correlation on ULP is suggestive but not statistically significant ($\rho = -0.43$, $p = 0.34$). Maverick (lowest fertility, highest ULP) and both Qwen models (highest fertility, below-average ULP) fit the expected pattern, but Nemotron Super~3 (moderate fertility, worst ULP) and Mistral Large~3 (moderate fertility, second-best ULP) break it.

This suggests that fertility is at best a \emph{weak} predictor of linguistic competence. A well-optimized tokenizer that preserves morphological boundaries may facilitate grammar tasks, but architecture and training data composition matter at least as much. Nemotron's Mamba-Transformer hybrid, optimized for long-range dependencies, appears to trade fine-grained morphological sensitivity for document-level reasoning---excelling at case outcome classification \citep[96.0\%]{ovcharov2026tokenizer} while failing on grammatical minutiae (27.7\% ULP).

\subsection{Practical Recommendations}

\begin{enumerate}[leftmargin=*]
    \item \textbf{Default to zero-shot} for morphologically rich languages. Validate few-shot per model and task before deploying. Consider chain-of-thought \citep{wei2022chain} as an alternative prompting strategy that does not require demonstration examples.
    \item \textbf{Start with tokenizer analysis.} Fertility rankings are domain-invariant; measure once, apply everywhere.
    \item \textbf{Ignore parameter counts} for non-English model selection. Use language-specific benchmarks.
    \item \textbf{Budget for the tokenizer tax.} A model with 1.6$\times$ higher fertility is 1.6$\times$ more expensive per document.
\end{enumerate}

\subsection{Mitigating the Tokenizer Tax}

Several strategies can reduce the disproportionate cost imposed on morphologically rich and underrepresented languages. We group them by the layer of the pipeline they target.

\paragraph{Vocabulary expansion.} The most direct remedy is to augment an existing BPE vocabulary with language-specific merge operations. \citet{rust2021good} demonstrated that tokenizer quality varies dramatically across languages and that fertility is a reliable proxy for downstream performance. \citet{zheng2021allocating} showed that reallocating vocabulary capacity -- adding merges for underserved scripts and morphemes -- reduces fertility by 15--30\% without degrading performance on high-resource languages. For Ukrainian, targeted addition of frequent morphological suffixes (e.g., \foreignlanguage{ukrainian}{-ння}, \foreignlanguage{ukrainian}{-ськ-}, \foreignlanguage{ukrainian}{-ість}) could close much of the gap with Polish and Czech.

\paragraph{Continued pre-training.} Language-adaptive pre-training on a domain-relevant corpus can improve subword representations even without modifying the vocabulary. \citet{gururangan2020don} showed that continued pre-training on domain and task data yields consistent gains; applied to a low-fertility language, it reduces the effective cost per token by improving per-token informativeness, partially compensating for the higher token count.

\paragraph{Tokenizer-free architectures.} Character-level and byte-level models eliminate the fertility disparity entirely. CANINE \citep{clark2022canine} operates directly on Unicode code points, while ByT5 \citep{xue2022byt5} processes raw UTF-8 bytes, achieving competitive performance without any learned vocabulary. The trade-off is computational: sequence lengths grow 3--5$\times$, increasing attention cost quadratically. These architectures are most attractive when equitable multilingual coverage outweighs inference budget constraints.

\paragraph{Tokenizer-aware inference.} When the model and its vocabulary are fixed -- as is typical with proprietary APIs -- practitioners can still mitigate the tax at inference time. Strategies include prompt compression, language-specific prompt templates that avoid high-fertility constructions, and cost-aware model routing that selects cheaper tokenizers for languages with elevated fertility. The fertility measurements reported in this paper (released as a public dataset) provide the empirical basis for such routing policies.

\section{Limitations}

\paragraph{SIB-200 test set size.} The test split contains only 204~examples with 7~classes, resulting in some classes having as few as 17~instances (geography). Confidence intervals on minority classes are wide.

\paragraph{Three domains.} Domain invariance of fertility is now tested on three registers (legal, news, encyclopedic), which strengthens the claim but still leaves specialized registers (biomedical, parliamentary, social media) untested. Similarly, the ``task-dependent'' few-shot conclusion rests on two tasks; a third classification task would increase confidence.

\paragraph{Slavic language coverage.} We test four Slavic languages. Including non-Slavic morphologically rich languages (e.g., Finnish, Hungarian, Turkish) would strengthen the morphological complexity hypothesis.

\paragraph{TMX segment quality.} EU Act translations may contain formulaic boilerplate that inflates cross-lingual similarity and deflates fertility differences.

\paragraph{ULP dataset size.} With 347~questions and only 7~data points in the fertility--competence correlation, statistical power is limited.

\paragraph{Model coverage.} We evaluate ten models, but the landscape evolves rapidly. Results may not generalize to models released after May~2026. Three models (Mistral Large~3, Nemotron, Nova~Pro) lack publicly available tokenizers and are measured only via API.

\section{Conclusion}

We presented a controlled cross-lingual tokenizer fertility map for ten models across 25~European languages, extended with subword decomposition analysis and mitigation strategies. Our key findings:

\begin{enumerate}[leftmargin=*]
    \item \textbf{The tokenizer tax spans 2.5$\times$} across European languages (English 1.23 to Greek/Maltese 3.1, mean across six local tokenizers). Ukrainian (2.66) pays 15--18\% more than cognate Slavic languages; corpus statistics confirm this reflects 2--6$\times$ less pre-training data than Polish despite comparable speaker populations.

    \item \textbf{Fertility is domain-invariant across three registers.} Model rankings are perfectly preserved ($\rho = 1.0$) between news and encyclopedic text, and consistent with legal text. A single measurement predicts cost across all domains.

    \item \textbf{Subword analysis reveals the mechanism.} High-fertility tokenizers fragment Ukrainian words at arbitrary byte boundaries rather than morpheme boundaries: Qwen3 splits ``\foreignlanguage{ukrainian}{відповідальність}'' into 9~subwords vs.\ 4 for Gemma~2, despite comparable vocabulary sizes. Vocabulary size alone does not predict efficiency; the proportion of Cyrillic-specific merges matters more.

    \item \textbf{Few-shot effects are model-intrinsic across languages.} Cross-lingual experiments on four Slavic languages show identical patterns: Maverick degrades everywhere ($-$8.4~pp avg), Nemotron improves everywhere ($+$10.5~pp avg).

    \item \textbf{Fertility weakly predicts grammatical competence} ($\rho = -0.43$, $n = 7$), but architecture matters more: Nemotron excels at classification while scoring worst on grammar.
\end{enumerate}

For practitioners: \textbf{measure fertility once} -- it predicts cost across all domains. \textbf{Validate few-shot per model and task}, not per language. \textbf{Budget for the Ukrainian tax}: 2.2$\times$ English cost per API call. Total experiment cost: \$4.05. All fertility measurements are released as a public dataset.

\section*{Acknowledgments}

This work was conducted as part of the LEX AI platform development at legal.org.ua. Compute costs for all experiments were covered by an AWS Activate grant. We thank the creators of SIB-200, the ULP benchmark, and the EU Acts in Ukrainian corpus for making their datasets publicly available.

\paragraph{Data and code availability.} All datasets used are publicly available: SIB-200 \citep{adelani2024sib200}, EU Acts in Ukrainian \citep{francophonicEU}, ULP \citep{galeshchuk2024ulp} -- all on HuggingFace. The cross-lingual fertility measurements and subword decomposition data are released as a public dataset at \url{https://huggingface.co/datasets/overthelex/tokenizer-fertility-map}. Experiment logs and measurement scripts are available at \url{https://github.com/overthelex/SecondLayer}.


\end{document}